\documentclass[conference]{IEEEtran}
\ifCLASSINFOpdf
\else
\fi
\usepackage{graphicx}
\usepackage{amsmath}
\usepackage{booktabs}
\usepackage{multicol} 
\usepackage{multirow} 
\usepackage{algorithm2e}
\usepackage{algpseudocode}
\usepackage{algpascal}
\usepackage{varwidth}
\usepackage{subcaption}
\usepackage[hang,flushmargin]{footmisc}
\usepackage{lipsum}
\usepackage[top=2.54cm, bottom=1.91cm, left=1.91cm, right=1.91cm]{geometry}

\begin{document}
\title{Novel Perception Algorithmic Framework For Object Identification \& Tracking In Autonomous Navigation}

\author{\IEEEauthorblockN{Suryansh Saxena}
\IEEEauthorblockA{National Robotics Engineering Center\\
Carnegie Mellon University\\
Pittsburgh, PA 15213\\
Email: ssaxena@nrec.ri.cmu.edu}
\and
\IEEEauthorblockN{Isaac K Isukapati}
\IEEEauthorblockA{The Robotics Institute\\
Carnegie Mellon University\\
Pittsburgh, PA 15213\\
Email: isaack@cs.cmu.edu}}


%Telephone: (800) 555--1212, Fax: (888) 555--1212}


\maketitle

\begin{abstract}
%\boldmath
This paper introduces a novel perception framework that has the ability to identify \& track objects in autonomous vehicle's field of view. The proposed algorithms don't require any training for achieving this goal. The framework makes use of ego-vehicle's pose estimation and a KD-Tree-based segmentation algorithm to generate object clusters. In turn, using a VFH technique, the geometry of each identified object cluster is translated into a multi-modal PDF and a motion model is initiated with every new object cluster for the purpose of robust spatio-temporal tracking. The methodology further uses statistical properties of high-dimensional probability density functions and Bayesian motion model estimates to identify \& track objects from frame to frame. The effectiveness of the methodology is tested on a KITTI dataset. The results show that the median tracking accuracy is around 91\% with an end-to-end computational time of 153 milliseconds.   
  
\end{abstract}


% \fi
\IEEEpeerreviewmaketitle

\section{Introduction}
\label{intro}
Perception systems play an instrumental role in the safe, successful, and reliable navigation of autonomous vehicles (AVs). Fundamentally, the perception system of an autonomous vehicle translates input data from the sensors into semantic information describing which objects are present, their associated pose, and the spatio-temporal relationship between them. Perception systems use segmentation \cite{wang2012could}, feature extraction \cite{azim2014layer}, and classification \cite{behley2013laser} techniques to generate this information. Traditional pipelines tend to optimize each technique individually. However, advancements in learning representations and deep learning methods \cite{ren2015faster, zhang2017towards} made it possible to develop end-to-end pipelines to optimize overall pipeline performance. Today, most perception research for AVs focuses on processing sensor data from cameras and LiDAR, primarily using 3D object detection methods. As pointed out by Arnold et al., \cite{arnold2019survey}, 3D object detection methods can be broadly divided into three categories: 1) monocular-image–based methods, 2) point-cloud–based (PCL) methods, and 3) fusion-based methods.

\textit{Monocular-image-based}  primarily focus on estimating 3D bounding boxes based on monocular images that lack information on depth perception. These methods predict a 3D bounding box for all identified 2D candidates. The bounding algorithms can be based on neural networks \cite{girshick2015fast}, geometrical constraints \cite{mousavian20173d}, or 3D model matching \cite{xiang2015data, chabot2017deep}.

\textit{Point-cloud-based} methods focus on processing the data produced by 3D scanners such as stereo cameras or LiDAR. Some PCL-based methods can be similar to monocular-image-based methods in the sense they first transform the PCL data into a 2D image using plane, spherical, or cylindrical project methods \cite{wu2018squeezeseg, su2015multi, li2016vehicle}, and then predict a 3D bounding box using spatiotemporal dimensional regressions. In related literature, these methods are typically referred to as projection-based methods. Other approaches such as volumetric convolution \cite{li20173d, engelcke2017vote3deep} or point-nets methods \cite{qi2017pointnet, bronstein2017geometric}are proposed to minimize the information loss.  

\textit{Fusion-based} methods make use of PCL data from 3D scanners and texture data from monocular images to improve the overall accuracy of the pipeline \cite{schlosser2016fusing}. 

While all these methods are extremely valuable for advancing the frontier of perception algorithms, they heavily rely on extensive and exhaustive training to achieve any reasonable level of performance. The training is data intensive, and large-scale image datasets like ImageNet \cite{deng2009imagenet}, KITTI \cite{geiger2012we}, and Virtual KITTI \cite{gaidon2016virtual} are specifically created for this purpose. Furthermore, it is practically impossible to exhaustively enumerate all possible real-world scenarios in a training dataset.

To address some of these limitations, we propose a novel perception algorithm—one that requires no training—for identifying \& tracking objects in an autonomous vehicle's field of view. According to the methodology, by using the pose-estimation information of an ego vehicle, the PCL data from a 3D scanner (stereo camera or LiDAR) is transformed into a world coordinate system. Then, a KD-Tree–based data segmentation algorithm (DBSCAN in this case) is used to associate a group of points with a particular obstacle. In turn, we employ a viewpoint feature histogram (VFH) technique to translate the geometry of each of the obstacle clusters into multi-modal probability density functions \cite{rusu2010fast}, and a motion model is initiated for each new object. These steps are repeated for each new data frame. 

Furthermore, the methodology uses the following three tests for the purpose of mapping object clusters from one frame to another. First, to identify feature similarity, we take the VFH associated with each object in the previous frame and run a chi-squared distance test on mean for all objects in the current frame. The purpose of this test is to generate a subset of possible matches for each object cluster. Second, we convert the VFH of each object and its potential matches into cumulative density functions (CDFs) by computing the volume integral of each cluster. We then employ a maximum deviation test (MDT) \cite{isukapati2016synthesizing, saxena2019multiagent} to compare statistical similarities between each of the candidate clusters and the subject object. The candidate with the highest MDT score is chosen as the final candidate. The MDT test, however, fails to identify scenarios in which objects with similar geometric shapes are in very close proximity to each other. To address this limitation, we employ a Bayesian motion model estimate that provides the likelihood of each candidate to reach that candidate object’s position in the light of new data. The candidate object with the highest likelihood is selected as a final candidate. A probability decay model is then employed to make decisions about cluster objects with no association. Finally, the usefulness of the proposed framework is evaluated using KITTI dataset.

%a chi-squared distance test is conducted between VFH of a specific object in previous frame and VFHs associated with every object in the current frame. The purpose of this test is to reduce the possible matches.        

%This information includes identification of both static (parked cars, traffic signs etc.), and dynamic objects (vehicles, bicyclists, pedestrians). Perception systems employ machine learning algorithms to identify \& track various agents in the environment. Without loss of generality, perception algorithms can be classified into two broad categories: 1) 2D object detection methods, and 2) 3D object detection methods.

The rest of the paper is organized as follows: Section \ref{concept} introduces the perception algorithmic framework in more detail, Section \ref{data} presents details of KITTI dataset used to test the proposed framework, Section \ref{analysis} presents analysis of the results, and Section \ref{conc}  provides conclusions and points out future lines of work.

\section{Perception Framework}
\label{concept}
\begin{figure}[!ht]
    \centering
    \includegraphics[width = 0.4\textwidth]{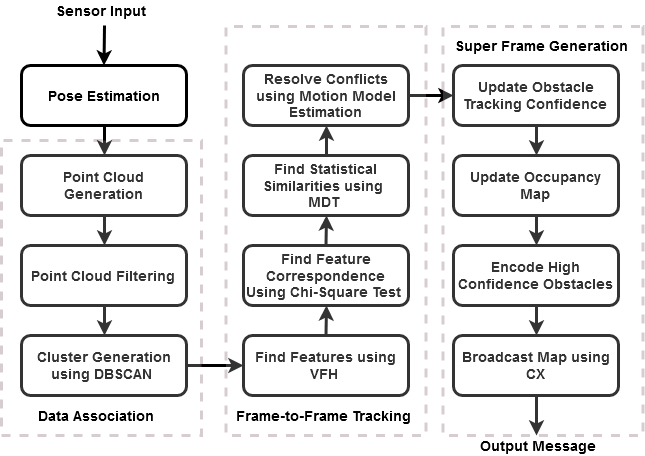}
    \caption{High Level Overview of Methodology}
    \label{fig:methodlogy_trac_state_machine}
\end{figure}

As mentioned in Section \ref{intro} the purpose of this paper is to present a novel framework for identifying \& tracking dynamic objects within the field of view of an autonomous vehicle. The framework doesn't require any training. A high level overview of the methodlody is presented in Figure \ref{fig:methodlogy_trac_state_machine}.

As it can be seen, the methodology can be divided into four steps: 1) Ego-vehicle pose estimation, 2) Data segmentation \& object identification, 3) Frame-to-frame object tracking, and 4) Super frame generation. Details on each of these steps are presented below.

\subsection{Ego-vehicle pose estimation}
An ego-vehicle must first understand its current location and orientation for the purpose of navigation. This is a critical task as it allows the ego-vehicle to identify and localize other objects within its field of view thereby ensuring the navigational safety. Typically, AVs make use of GPS, velocity, and IMU data for this task. In this paper, we employ EKF algorithm for this task. The algorithm uses IMU data for belief state estimation, and GPS \& velocity data for correcting measurement errors. The output of the algorithm is a 6D pose estimation of the ego-vehicle that includes its location and orientation.  

\subsection{Data filtering, segmentation \& object identification}

Given our main objective in this framework is to identify and track only static \& dynamic agents in the environment, the segments of LiDAR point cloud that represent the ground plane are inconsequential for our purposes. So, to filter out these points, we model the point cloud data pertaining to the ground/pavement to lie along the plane 'ax + by + ca + d = 0', and make use of RANSAC algorithm to filter out these points \cite{fischler1981random}. Next, using the pose estimation information of an ego-vehicle generated in the previous step, the filtered PCL data from a 3D scanner (LiDAR in this case) is transformed into a world coordinate system. 

A KD-Tree based data segmentation algorithm (DBSCAN in this case) is then used to associate a group of points with a particular obstacle \cite{borah2004improved}. The computational complexity of this algorithm is \textit{O(n log n)}. In turn, VFH technique in employed to translate the geometry of each of the obstacle clusters into multi-modal probability density functions.  A motion model with seven parameters [x, y, z, $v_x$, $v_y$, $\theta$] is initiated for each new object. At the outset, all velocity values in the state vector are set to zero. Steps 1 and 2 are repeated for each new data frame.

%The following three steps are taken to accomplish this objective: First, we limit the circle of influence of the ego-vehicle to a 30 meter radius. While it is true that the radius of the field of view of a typical LiDAR scan is 120 meters, point cloud data for the objects farther away is very noisy. and hence, our algorithm filters out any data outside the circle of influence. Second, in most scenarios, many points from the LiDAR beam fall on the ground. To filter out these points, we model the point cloud data pertaining to the ground/pavement to lie along the plane 'ax + by + ca + d = 0', and make use of RANSAC algorithm to filter out these points.

\subsection{Frame-to-frame object tracking}

Frame-to-frame object tracking is the most critical step of this pipeline. As the reader might know, the ability to quantify and resolve uncertainties in object mapping is non-trivial. We carefully considered computational complexity and statistical robustness while trying to minimize the mismatches.

\RestyleAlgo{boxruled}
\alglanguage{pseudocode}
\begin{algorithm}
	\SetAlgoLined
	%\KwData{raw data}
	%\KwResult{composite CDF}
	$A_{n-1}$ = set of object clusters in frame 'n-1'\;
	$A_n$ = set of object clusters in frame 'n'\;
	$a_{n-1}^i$ = object cluster 'i' in $A_{n-1}$\;
	$a_{n}^j$ = object cluster 'j' in $A_{n}$\;
	$P_{(i,j)}$ = subset of potential matches of $a_{n-1}^i$ in $A_n$\;
	$F_{n-1}^i$, $F_{n}^j$ = CDFs of $a_{n-1}^i$, $a_{n}^j$\;
	$S_{(i,j)}$ = vector of MDT scores\;
	
		\For {$a_{n-1}^i$ $\in$ $A_{n-1}$}
		{\For {$a_{n}^j$ $\in$ $A_{n}$}
		{perform $\tilde{\chi}^2$ distance test between [$a_{n-1}^i$, $a_{n}^j$]\;
		\If {$\tilde{\chi}^2$ == True}
		{add $a_{n}^j$ to $P_{(i,j)}$}
		}
		\If {$n(P_{(i,j)})$ $>$ 1}
		{\For {$a_{(i,j)}$ $\in$ $P_{(i,j)}$}
		{compute $s_{i,j}$ between $F_{i,j}$ and $F_{n-1}^i$\;
		update $S_{(i,j)}$}
		}
		
		\If {MDT scores tied == True}
		{use motion model to pick final candidate}
		\Else{pick cluster with highest MDT as a match}
		
		\If {$n(P_{(i,j)})$ == 1}
		{$a_{n}^j$ $\in$ $n(P_{(i,j)})$ is the match}
		
		\If {$P_{(i,j)}$  = = $\emptyset$}
		{obstacle match not found}
		Update the confidence using decay model
		
		} 		 
	\caption{Frame-to-frame object tracking}
\end{algorithm}

Algorithm 1 describes methodology for frame-to-frame object tracking. As described in the algorithm the sets $A_{n-1}$ and $A_n$ contain object clusters in frames 'n-1', and 'n' respectively. Let $a_{n-1}^i$, and $a_{n}^j$ respectively represent object clusters 'i' and 'j' in $A_{n-1}$ and $A_n$. Let $F_{n-1}^i$ and $F_{n}^j$ represent CDFs associated with object clusters $a_{n-1}^i$ and $a_{n}^j$ respectively. Finally, $P_{(i,j)}$ represents a subset of potential matches for the object $a_{n-1}^i$ in $A_n$. 

The algorithm makes use of up to three different tests while finalizing the cluster mapping. For each $a_{n-1}^i$ in $A_{n-1}$ a chi-squared distance test is conducted between $a_{n-1}^i$ and with every member of $A_n$. Every $a_{n}^j$ that passes the test is considered a potential match and is added to the set $P_{(i,j)}$. On the other hand, if the set $P_{(i,j)}$ has only one member, then that member is automatically treated as a match. If the set $P_{(i,j)}$ is an empty set, the object cluster is flagged for no match and the confidence of algorithm for tracking this obstacle is decreased using a probability decay model. If the confidence of tracking will be increased if a match is found in the subsequent frame and so on. At any point in time, if the confidence of tracking an object cluster falls below a threshold value, then that cluster will be discarded. Lastly, if the set $P_{(i,j)}$ has more than one potential match, we score the statistical similarity between every $F_{i,j}$ in $P_{(i,j)}$ and $F_{n-1}^i$ (please note that these cumulative density fuctions are generated by computing the volumetric integral of the corresponding VFH). The object cluster with highest MDT score is picked as a final match. We make use of a motion model estimate to resolve ties in MDT scores for the candidates. Equations \ref{eq:StateVector}, \ref{eg:Vel}, \ref{eg:AccT}, \ref{eg:AccN} represent the state equations for the motion model. We make use of EKF updates in the Bayesian setting to generate motion model updates. Please note since this test requires a motion estimate, it is only possible after the third frame from the initialization.

\begin{subequations}
\label{eq:Tracking}
\begin{align}
         SV = [X,Y,Z,V_x,V_y,\theta],  \label{eq:StateVector} \\
         V  = \frac{X_{n} - X_{n-1}}{\partial T},  \label{eg:Vel} \\
         A_T  = \frac{\partial V}{\partial T}, \label{eg:AccT}\\
         A_N  = \sqrt{A - A_T} \label{eg:AccN}
\end{align}
\end{subequations}

\RestyleAlgo{boxruled}
\begin{algorithm}[!ht]
\SetKwBlock{kwInit}{Initialize Vehicle $(av_i)$}{end}
\caption{Data Analysis Flow}\label{algo_1_methodology}
\KwData{Vehicle sensor configuration $( S_i)$}
\KwData{Initial vehicle pose estimate $(x_i)$}
\kwInit{
        \par
        \vspace{0.1cm}
        \Indp{
           initiate occupancy map $(M_i)$\;
           initiate tracked obstacle vector $({Obs}_i)$\;
           initiate raw sensor data vector $({S}_i)$\;
           set point cloud generated  $(c_i = 0)$\;
        }
        \vspace{0.1cm}
        }

 \While{ (${av}_i$ is alive) }{
     ${S_i}^j = $ \Call{Read Sensor Input}{j}, $j \in Sensors(\S_i)$\; 
     $(x_i) = $ \Call{Update Pose}{$S_i , x_i$}\;
     Point Cloud $(p_i) = $ \Call{Generate Point Cloud}{$x_i, S_i$}\;
     $c_i = c_i + 1 $\;
     $(p_i) = $ \Call{Filter Point Cloud}{$p_i$}\;
     $(p_i) = $ \Call{Remove Ground Plane}{$p_i$}\;
     Object Cluster $ (nObs_i) =$ \Call{DBSCAN}{$p_i$}\;
     ${nObs_i}^k \xrightarrow{} \Call{Compute VFH}{} , k \in nObs_i$\;
     $\Call{Frame To Frame Object Tracking}$\;
     ${Obs_i}^j = \Call{Update Vector}{{nObs_i}^j,\alpha_i}, j \in nObs_i$\;
    ${kdTree}_i = \Call{Update KdTree}{{Obs_i}^j}, j \in Obs_i$\;
        $ \Call{Update Occupancy Map}{M_i, Obs_i}$\;
        \If{$Time(t) \geq Trigger Time$}
            { $Super Frame( {SF}_i) \xrightarrow{} initialize$\;
              $O_i = \Call{GetHighConfidenceClusters}{M_i}$\;
              $\Call{Broadcast Msg}{Encode Msg(O_i)}$\;
            }
}
\end{algorithm}

\subsection{Super frame generation}
At the end of the previous step, the occupancy map for every existing object is registered onto an Octomap and VFH and motion model are initiated for any newly identified objects. As the algorithm marches forth in time, the confidence of tracking of objects in ego-vehicle's environment increases. Even though, it is not within the scope of this paper, the information from the Octomap can be encoded and shared with other AVs in the vicinity for other multi-agent coordination applications.

\section{Data set}
\label{data}
\begin{figure}[!ht]
    \centering
    \includegraphics[width = 0.5\textwidth]{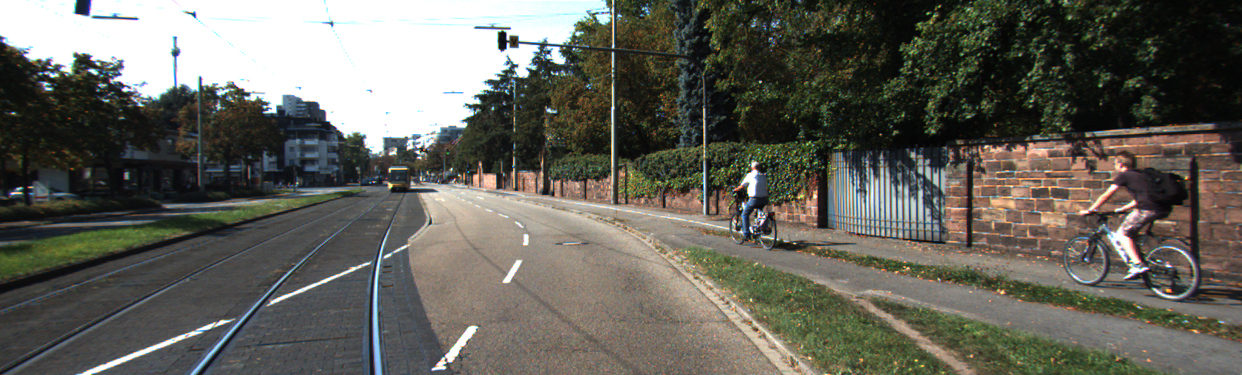}
    \caption{Schematic of the data set}
    \label{fig:schematic}
\end{figure}

We tested the framework described in Section \ref{concept} using a KITTI dataset. The dataset we used consists of PCL data from both stereo camera and LiDAR, GPS data, IMU data, IMU to LiDAR calibration matrix, and local transforms. We used only LiDAR PCL data in our experiments. As shown in Figure \ref{fig:schematic} the selected data set consists of 2 bicyclists traveling in the same direction as that of the ego-vehicle, and another automobile traveling in the opposite direction of ego-vehicle. We developed the software infrastructure for both topic generation and perception framework implementation. Algorithm 2 provides high-level details of the data analysis flow.    

\vspace{-0.3cm}

%\vspace{-0.3cm}

\section{Analysis of results}
\label{analysis}
The analysis section is divided into three subsections: first, we will make some general observations on analysis; second, we will discuss details of a scenario where the proposed algorithm was able to distinguish two bicyclists next to each other; third, we will present tracking accuracy statistics of the framework, and finally we will discuss computational scalabity of the framework. 

\subsection{General observations}
\begin{figure}[!ht]
\centering
%\begin{center}
%\hspace*{-18mm}%
\includegraphics[width = 0.5\textwidth]{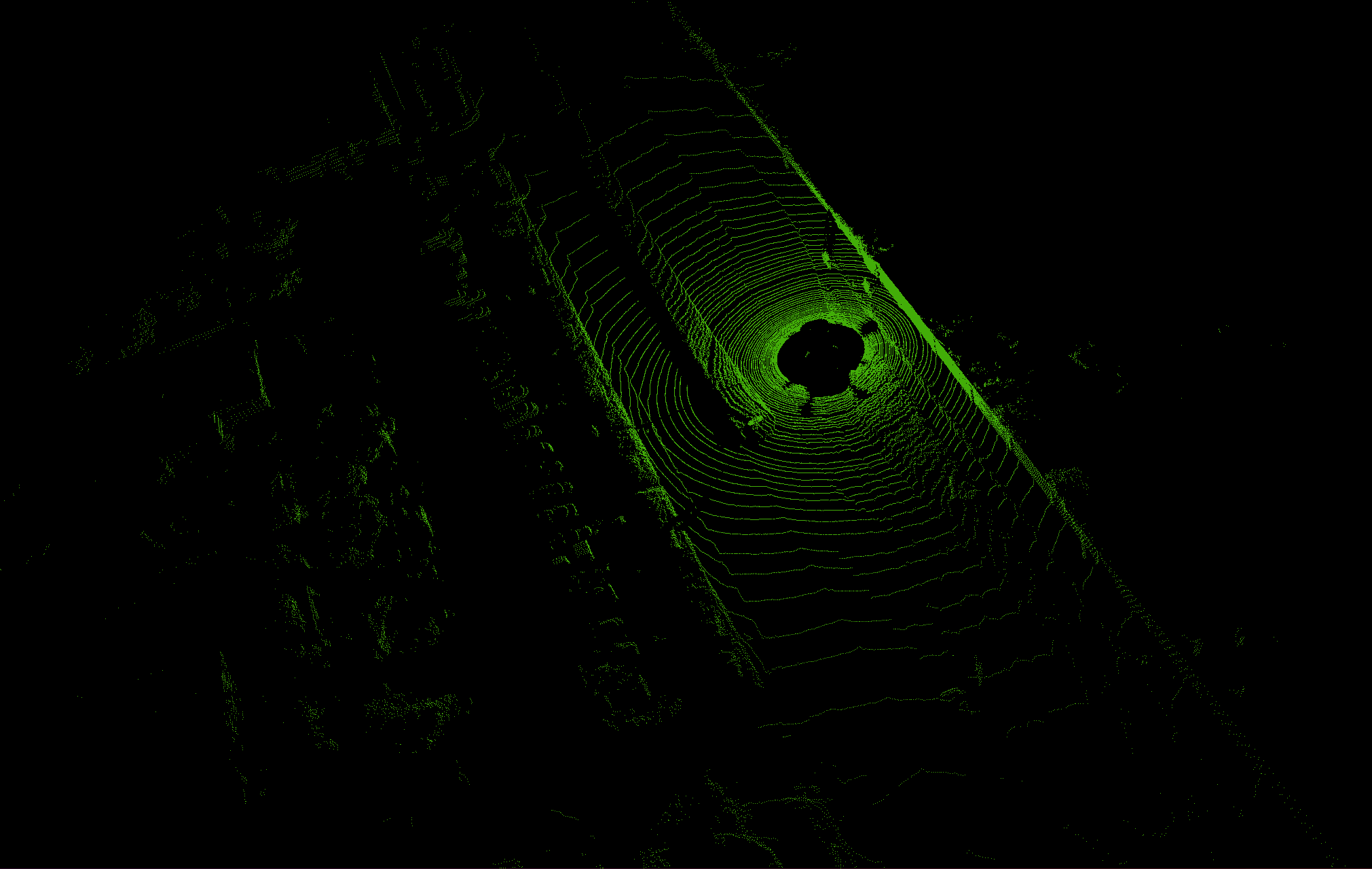}
%\end{center}
\caption{LiDAR Point Cloud data)}
\label{fig:pcl}
\end{figure}

\begin{figure}[!ht]
\centering
%\begin{center}
%\hspace*{-18mm}%
\includegraphics[width = 0.5\textwidth]{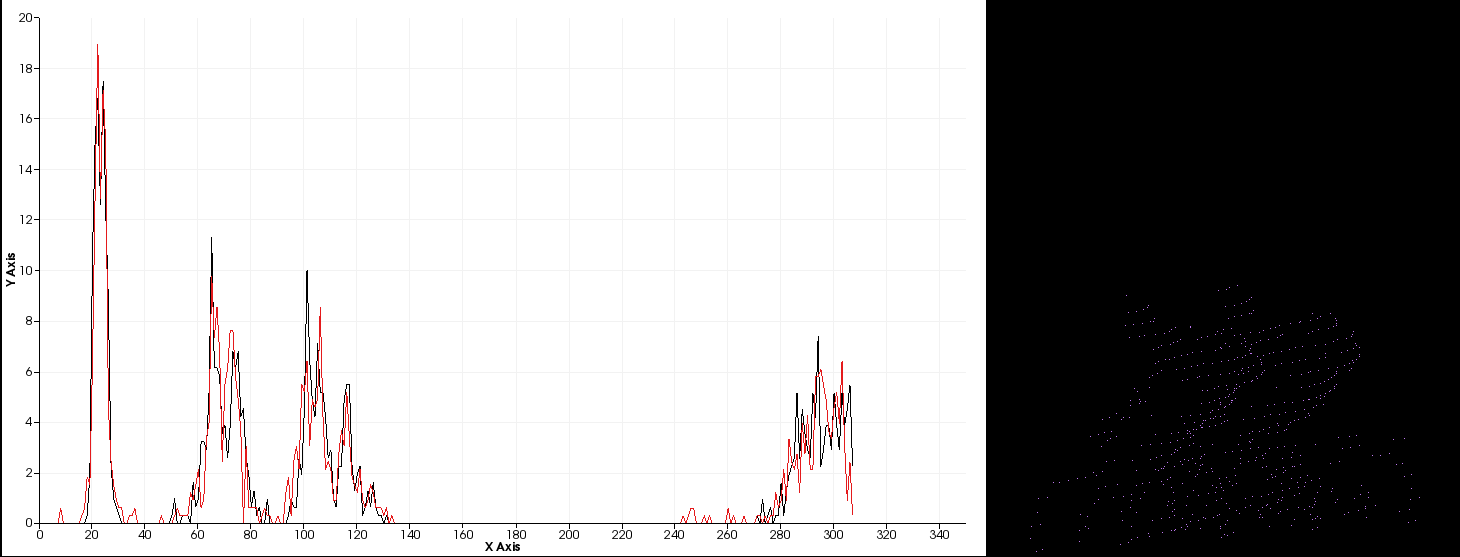}
%\end{center}
\caption{Example VFH output)}
\label{fig:vfh}
\end{figure}

\begin{figure}[!ht]
\centering
%\begin{center}
%\hspace*{-18mm}%
\includegraphics[width = 0.5\textwidth]{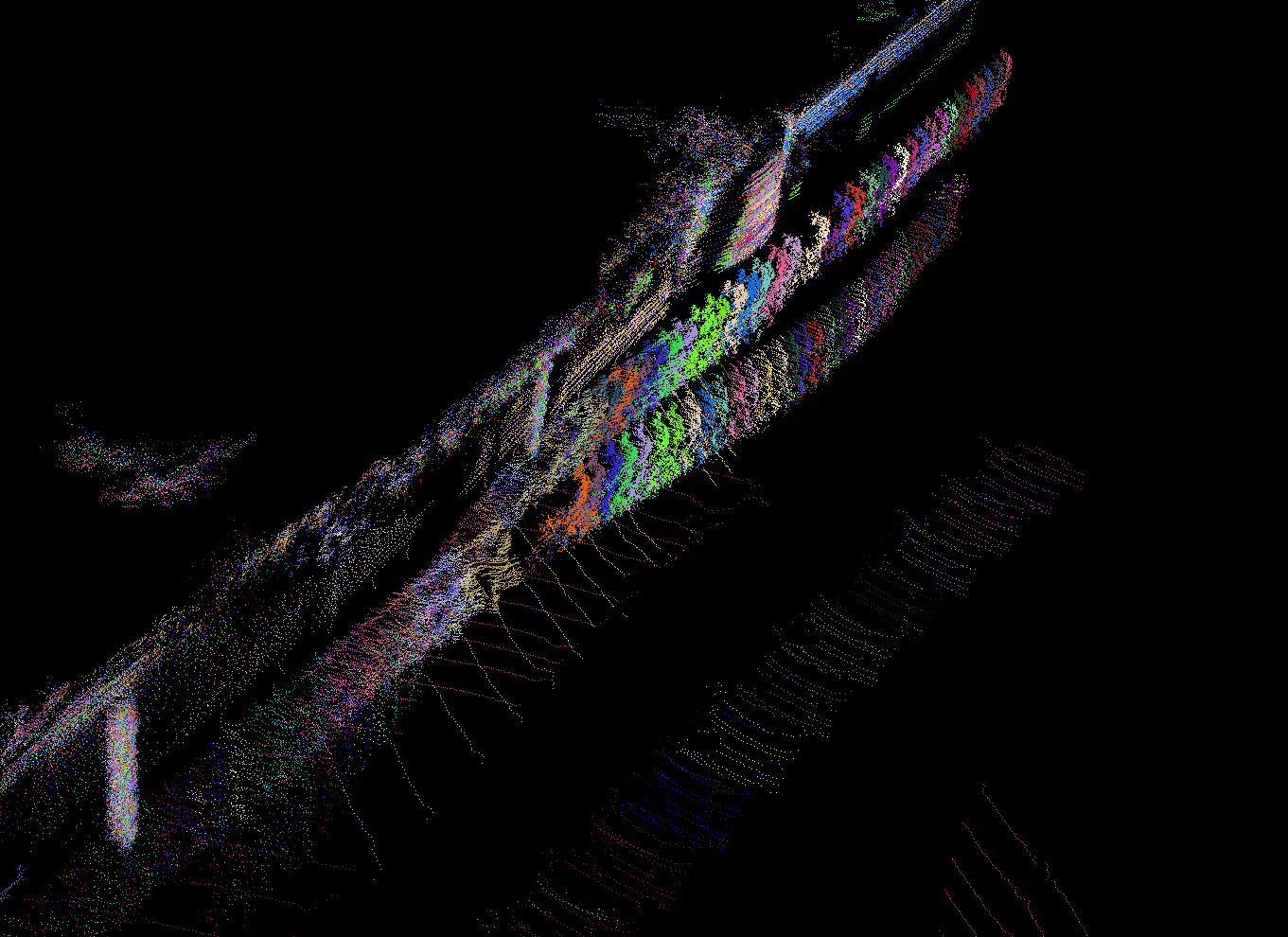}
%\end{center}
\caption{Tracking across frame)}
\label{fig:output}
\end{figure}

Figure \ref{fig:pcl} presents a schematic of the raw LiDAR point cloud data provided by the KITTI data set. Figure \ref{fig:vfh} represents viewpoint feature histogram of two identified objects. The x-axis of this figure represents the coded geometric feature, whereas the y-axis represents its corresponding frequency. As it is evident from the figure these histograms are high dimensional in nature. Lastly, Figure \ref{fig:output} represents frame to frame output generated by our perception algorithmic framework.    

\subsection{Ability to resolve data association issues}
\begin{figure}[!ht]
\centering
%\begin{center}
%\hspace*{-18mm}%
\includegraphics[width = 0.5\textwidth]{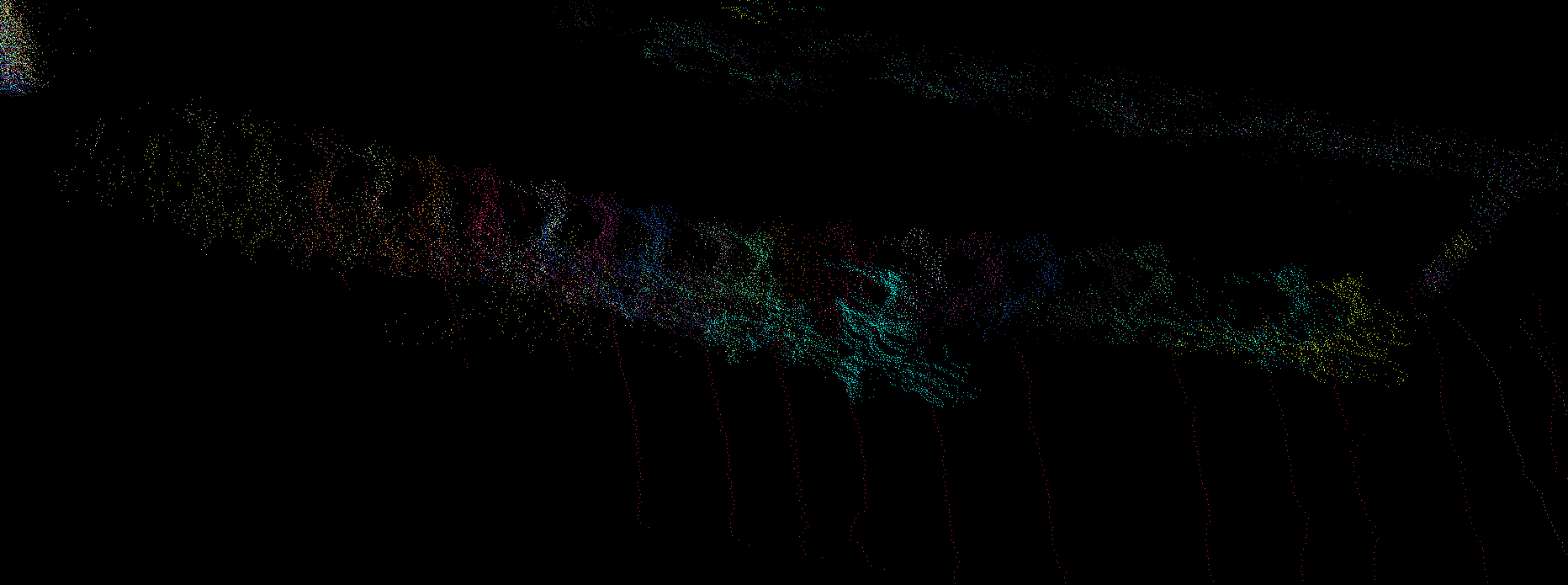}
%\end{center}
\caption{Data association issue}
\label{fig:issue}
\end{figure}

\begin{figure}[!ht]
\centering
%\begin{center}
%\hspace*{-18mm}%
\includegraphics[width = 0.5\textwidth]{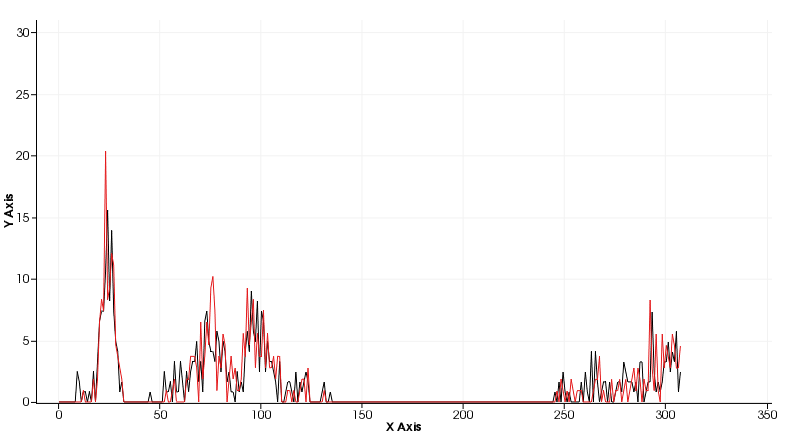}
%\end{center}
\caption{Failed MDT test}
\label{fig:mdt-fail}
\end{figure}

\begin{figure}[!ht]
\centering
%\begin{center}
%\hspace*{-18mm}%
\includegraphics[width = 0.5\textwidth]{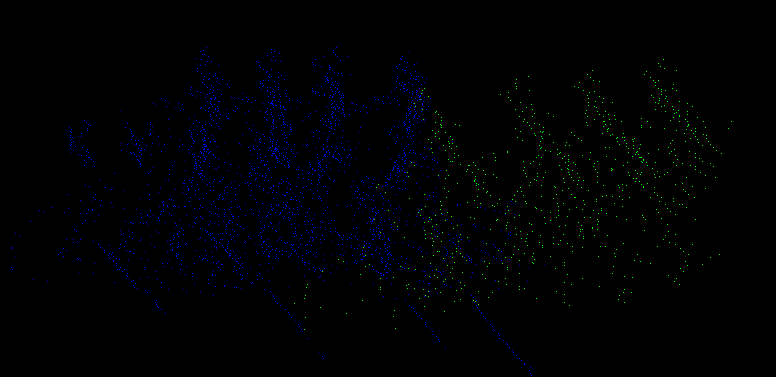}
%\end{center}
\caption{Demonstration of successful tracking}
\label{fig:success-frame}
\end{figure}

As shown in Figure \ref{fig:output} and Figure \ref{fig:issue} for a part of the duration the bicyclists were so close to each other such that their corresponding viewpoint feature histograms are statistically significantly similar. These VFHs are presented in Figure \ref{fig:mdt-fail}. As a result maximum deviation test failed to distinguish the CDFs associated with these VFHs as two distinct objects. However, as demonstrated in Figure \ref{fig:success-frame} the motion model estimate was able to distinguish that these object clusters belong to two different objects. This scenario demonstrates the robustness of the proposed framework in object identification \& tracking.  

\subsection{Tracking accuracy}
\begin{figure}[!ht]
\centering
%\begin{center}
%\hspace*{-18mm}%
\includegraphics[width = 0.5\textwidth]{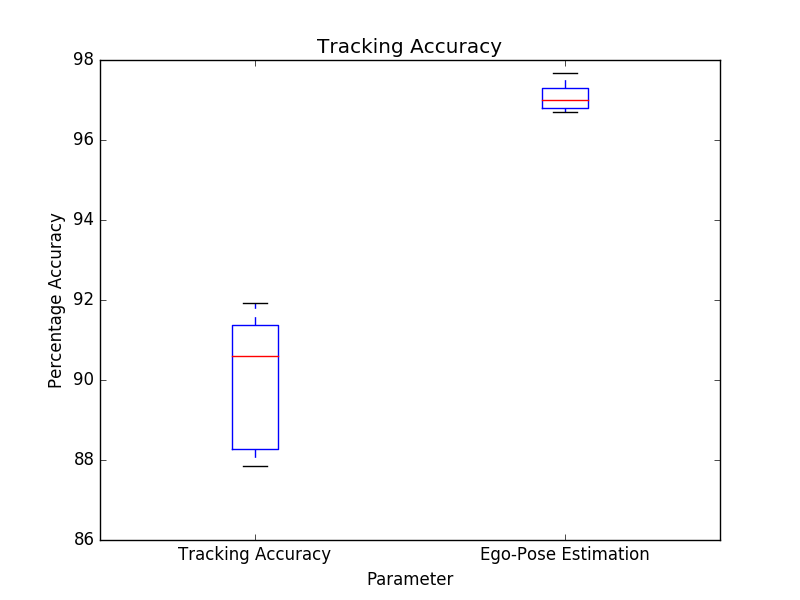}
%\end{center}
\caption{Boxplot of tracking accuracy}
\label{fig:statistics}
\end{figure}

To summarize the usefulness of the proposed framework, we computed summary statistics for tracking accuracy and ego-pose estimation. The figure \ref{fig:statistics} presents a standard boxplot for summarizing the descriptive statistics of these parameters: red circles  in  the plot represent median  success values, whereas the values within  the  box  represent  the  data  within  the inter-quartile range. As it can be seen, the tracking accuracy of the framework ranged between 87\% - 92\% with the median around 91\%. Similarly, the ego-pose estimation has a median accuracy of around 97\%.   

\subsection{Computational times}
We logged the process times of various processes in order to evaluate the computational scalability of the proposed algorithms. The tests are conducted on a machine with a 2-core Intel i7 2.7 GHz processor. The average statistics of the computational times are presented in table \ref{table:times}. As it can be seen, the table has two columns: column-1 provides information on the process name, whereas column-2 contains information on the average computational time in milliseconds. The average computational time of the overall methodology is about 153 milliseconds or 6.5Hz. Please note that this time is without employing any code optimization techniques. Furthermore, data association and PCL filtering take about 86\% of the computational times. We are in the process of updating these algorithms. 

\begin{table}[!ht]
\begin{tabular}{|c|c|}
\hline
\textbf{PROCESS} & \textbf{COMPUTATION TIME (ms)} \\ \hline
LiDAR Point Cloud Filtering & 57.42 \\ \hline
Pose Estimation (EKF) & 15.87 \\ \hline
Point Cloud Transformation & 13.23 \\ \hline
Data Association (DBSCAN) & 75.49 \\ \hline
Frame-to-frame mapping (MDT) & 19.25 \\ \hline
Methodology Total & \textbf{153.4615 (6.5 Hz)} \\ \hline
\end{tabular}
\caption{Processing Times on 2-Core Intel i7 2.7 GHz Processor}
\label{table:times}
\end{table}

\section{Conclusions and Future Work}
\label{conc}

In this paper, we introduced a novel perception framework that has the ability to identify \& track objects in autonomus vehicle's field of view. The algorithmic framework doesn't require any training for achieving this goal. It makes use of ego-vehicle's pose estimation and a KD-Tree-based DBSCAN algorithm to generate object clusters. The geometry of each object cluster is translated into a multi-modal PDF using a VFH technique. An associated motion model is initiated with every new object cluster for the purpose of robust spatio-temporal tracking of that object. The methodology further uses statistical properties of high-dimensional probabilistic functional spaces and Bayesian motion model estimates to identify \& track objects from frame to frame. The effectiveness of the methodology is tested on a KITTI dataset. The results demonstrate the robustness of algorithmic framework both in terms of tracking accuracy and computational scalability. 

In future, we intend to extend the framework in the following directions:
First, we want to employ code optimization techniques to further enhance the computational scalability. Second, we are interested in fusing the proposed algorithms with our multi-agent sensor fusion algorithms for the purpose of identifying and resolving false negatives in autonomous navigation \cite{saxena2019multiagent}. Third, we are interested in developing computationally scalable physics-based simulation infrastructure for testing these ideas further.

%\vspace{-5cm}





\bibliographystyle{IEEEtran}
\bibliography{ieee-2015.bib}
%\begin{}{1}

%\end{thebibliography}

%\end{thebibliography}

\end{document}